\documentclass[letterpaper]{article} 
\usepackage{aaai24}  
\usepackage{times}  
\usepackage{helvet}  
\usepackage{courier}  
\usepackage[hyphens]{url}  
\usepackage{graphicx} 
\usepackage{natbib}  
\usepackage{caption} 
\usepackage{algorithm}
\usepackage{algorithmic}

\usepackage{color} 
\usepackage{soul} 

\usepackage{epsfig}
\usepackage{amsmath}

\DeclareMathOperator{\mse}{mse}

\usepackage{amssymb}
\usepackage{booktabs, multirow}
\usepackage{xcolor}

\usepackage{paralist}

\newcommand{\calC}{{\cal C}}

\usepackage{newfloat}
\usepackage{listings}
\DeclareCaptionStyle{ruled}{labelfont=normalfont,labelsep=colon,strut=off} 
\floatstyle{ruled}
\newfloat{listing}{tb}{lst}{}
\floatname{listing}{Listing}
\nocopyright 

\title{Multi-View People Detection in Large Scenes via Supervised  View-Wise \protect \\ Contribution Weighting}
\author{
    Qi Zhang\textsuperscript{\rm 1}, Yunfei Gong\textsuperscript{\rm 1}, Daijie Chen\textsuperscript{\rm 2,\rm 1}, Antoni B. Chan\textsuperscript{\rm 3}, Hui Huang\textsuperscript{\rm 1}\footnote{Corresponding author.}
}
\affiliations{
    \textsuperscript{\rm 1}College of Computer Science and Software Engineering, Shenzhen University, Shenzhen, China\\
    \textsuperscript{\rm 2}Guangdong Laboratory of Artificial Intelligence and Digital Economy (SZ), Shenzhen, China \\
    \textsuperscript{\rm 3}Department of Computer Science, City University of Hong Kong, Hong Kong SAR, China\\
    qi.zhang.opt@gmail.com, gongyunfei2021@email.szu.edu.cn, chendaijie2022@email.szu.edu.cn,\\ abchan@cityu.edu.hk, hhzhiyan@gmail.com
}

\usepackage{bibentry}

\begin{document}
\maketitle

\begin{abstract}
  Recent deep learning-based multi-view people detection (MVD) methods have shown promising results on existing datasets. However, current methods are mainly trained and evaluated on small, single scenes with a limited number of multi-view frames and fixed camera views. As a result, these methods may not be practical for detecting people in larger, more complex scenes with severe occlusions and camera calibration errors. This paper focuses on improving multi-view people detection by developing a supervised view-wise contribution weighting approach that better fuses multi-camera information under large scenes. Besides, a large synthetic dataset is adopted to enhance the model's generalization ability and enable more practical evaluation and comparison. The model's performance on new testing scenes is further improved with a simple domain adaptation technique. Experimental results demonstrate the effectiveness of our approach in achieving promising cross-scene multi-view people detection performance. See code here: https://vcc.tech/research/2024/MVD.
\end{abstract}

\section{Introduction}
Multi-view people detection (MVD) has been studied to detect people's locations on the ground of the scenes via synchronized and calibrated multi-cameras, which could be used for many different applications, such as public safety, autonomous driving, \emph{etc}. Recent multi-view people detection methods are mainly based on deep learning, which train convolution neural networks (CNNs) with synchronized multi-view images as input and ground-plane occupancy map as output, and have achieved promising results on existing datasets, such as Wildtrack \cite{chavdarova2018wildtrack} and MultiviewX \cite{hou2020multiview}.

\begin{figure}[t]
\begin{center}
   \includegraphics[width=0.95\linewidth]{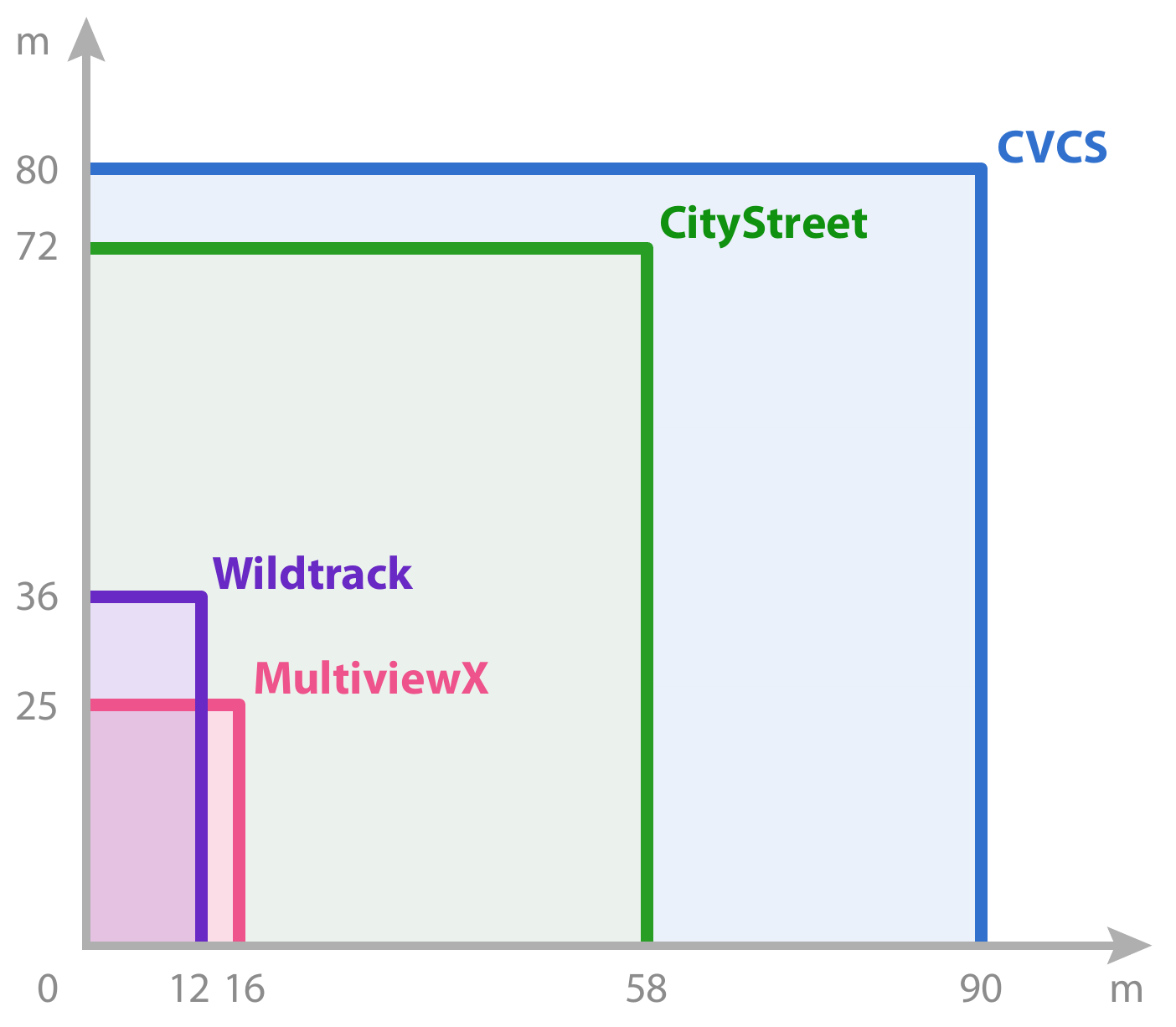}
\end{center}
   \caption{The scene area comparison of CVCS, CityStreet, Wildtrack, and MultiviewX.
   The scene size of the latter two datasets is quite smaller than the first two.}
\label{fig:scene_area}
\end{figure}

However, the current DNNs-based multi-view people detection methods are trained and evaluated only on single small scenes (see Figure~\ref{fig:scene_area}) with limited numbers of frames and fixed camera views. These datasets are collected on small scenes with only hundreds of frames for training and testing and several fixed camera views (7 in Wildtrack and 6 in MultiviewX). In summary, the weaknesses of current methods are 3 folds:
1) \emph{The methods are evaluated on small scenes} (about $20m*20m$), while real-world scenes could be much larger, with more severe occlusions and
camera calibration errors;
2) \emph{The methods are evaluated on datasets containing limited frames and fixed camera views} (\emph{e.g.}, 360 for training and 40 for testing, and 7 views in Wildtrack dataset), which could not validate and compare different methods thoroughly;
3) \emph{The methods cannot generalize to other scenes well} since they are trained on the same single scenes, and potentially overfitted on the specific camera arrangement, making them not generalized to novel scenes and camera layouts.
These settings in current multi-view detection methods should be adjusted to better validate and compare different multi-view detection methods.

\begin{figure*}[t]
\begin{center}
   \includegraphics[width=0.94\linewidth]{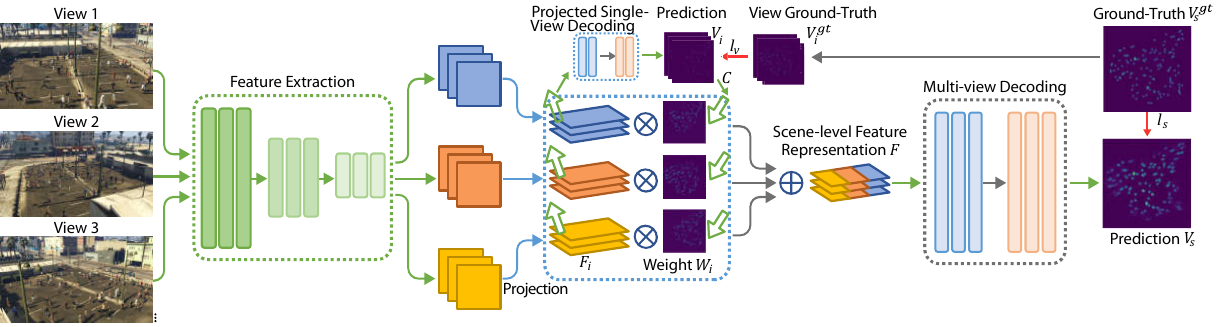}
\end{center}
   \caption{The pipeline of the proposed view-wise contribution weighting method, which consists of 4 stages:
   \emph{Single-view feature extraction and projection}, \emph{Projected single-view decoding},
    \emph{Supervised view-wise contribution weighted fusion}, and \emph{Multi-view feature decoding}.
    First, camera view features are extracted from the shared feature extraction net, and then they are projected to the ground plane.
    Second, each view's projected feature $F_i$ is fed into a decoder to predict the view's people location map $V_i$ on the ground,
    and the loss is $\ell_v$, whose ground-truth is obtained from the scene ground-truth $V^{gt}_s$.
    Third, each view's people location map prediction $V_i$ is fed into a subnet $\calC$ and then weighted across all camera views to
    obtain weight maps $W_i$ for multi-view fusion. And the predicted weight maps $W_i$ are used to fuse multi-view features $F_i$ in a weighted
    summation way. Finally, the fused multi-view feature $F$ is decoded to predict the whole scene's people location map, and the loss is $\ell_s$.}
\label{fig:pipeline}
\end{figure*}

In this paper, we focus on the multi-view people detection task in large scenes (eg. CVCS and CityStreet, see Figure~\ref{fig:scene_area}) with more occlusions and camera calibration errors, as well as the model's generalization ability to novel unseen scenes in testing.
We propose the supervised view-wise contribution weighting method to fuse multi-camera information on the scene ground plane based on each view's prediction on the
ground plane space. As shown in Figure~\ref{fig:pipeline}, the proposed supervised view-wise contribution weighting MVD model consists of 4 stages:
\emph{Single-view feature extraction and projection}, \emph{Projected single-view decoding}, \emph{Supervised view-wise contribution weighted fusion}, and \emph{Multi-view feature decoding}.
First, the features of each view are extracted and then projected to the ground plane in a shared subnet to handle possible different numbers of camera views.
The projected single-view decoding subnet predicts each view's people location map contained by the view on the ground plane in a supervised way, which could be used as the contribution of the current view to the final result. Thus, the predictions are further fed into a subnet and weighted across all camera views to obtain weight maps for multi-view fusion in the next step. Then the predicted weight maps are used to fuse multi-view features in a weighted summation way. Finally, the fused multi-view features are decoded to predict the whole scene's people location map.

Besides, in the experiments, instead of evaluating the multi-view people detection methods on small multi-view datasets, we adopt 2 large multi-view datasets, CitySteet \cite{zhang2019wide} and CVCS \cite{zhang2021cross}, for a more challenging and thorough method comparison and validation. Furthermore, a simple domain adaptation technique is also adopted to further improve the model's cross-scene performance on testing scenes. In summary, the main contributions of our paper are as follows.

\begin{itemize}
  \item To our knowledge, this is the first study on large-scene multi-view people detection task with better generalization ability to novel unseen
  testing scenes with different camera layouts.
  \item We propose a new multi-view people detection method, which considers the supervised view-wise contribution weighting for better multi-view feature fusion.
  \item The proposed method's cross-scene multi-view people detection performance is promising compared to previous methods trained on the same single scenes, extending multi-view people detection to more practical scenarios.
\end{itemize}

\section{Related Work}

\subsection{Multi-View People Detection}
\textbf{Traditional methods.}
Multi-view people detection has been studied for dealing with heavy occlusions in crowded scenes.
Usually, information from synchronized and calibrated multi-camera views is combined to provide predictions for the whole scene.
Early detection methods try to detect each person in the images by extracting hand-crafted
features \cite{viola2004robust, sabzmeydani2007detecting, wu2007detection} and then training
a classifier \cite{joachims1998text, viola2005detecting, gall2011hough} using the extracted features.
\citet{fleuret2007multicamera} proposed the Probabilistic Occupancy Map (POM) to indicate the probability of people appearing
on the grid of the scene ground.
Traditional methods rely on hand-crafted features and background subtraction preprocessing, which limit their performance and application scenarios.

\textbf{Deep learning methods.}
With the development of geometric deep learning, recent learning-based MVD methods have achieved great progress.
\citet{chavdarova2017deep} proposed to use CNNs for feature extraction and concatenate multi-view features to predict the occupancy map.
However, the features from different camera views are not aligned before further fusion in the model, resulting in limited performance.
\citet{hou2020multiview} used camera calibrations to perform a perspective transformation to the ground for feature fusion and achieved
state-of-the-art performance. Later work \citet{song2021stacked} improved the performance further by using multi-height projection with an attention-based soft selection module for different height projection fusion.
\citet{hou2021multiview} adopted the deformable transformer framework \cite{zhu2020deformable} and proposed a
multi-head self-attention based multi-view fusion method.
\citet{qiu20223d} proposed a data augmentation method by generating random 3D cylinder occlusions on the ground plane to relieve model overfitting.

Overall, the existing multi-view people detection methods are trained and evaluated on single small scenes with only hundreds of
multi-view frames and several fixed camera views, such as in Wildtrack \cite{chavdarova2018wildtrack} and MultiviewX \cite{hou2020multiview}.
This is not suitable for better validating and comparing different multi-view people detection methods, not to mention for generalizing to novel new scenes with different camera layouts, or other more practical real-world application scenarios.
\citet{qiu20223d} noticed the issue and tried to solve the problem from the aspect of data augmentation, but still evaluated the methods only on small scenes.
\emph{Besides, in contrast to SHOT \cite{song2021stacked} or MVDeTr \cite{hou2021multiview} which uses self-attention weights, the proposed method estimates the view fusion weights in a supervised way without extra labeling efforts, resulting in more stable performance.}

\subsection{Other Multi-View Vision Tasks}

\textbf{Multi-view counting}. Multi-camera views can be combined to further improve the single-image counting \cite{cheng2019learning,huang2020stacked,zhang2022crossnet,cheng2022rethinking,cheng2019improving} performance for large scenes.
Similar to multi-view people detection, traditional multi-view counting methods also rely on hand-crafted features and
background subtraction techniques \cite{viola2004robust,sabzmeydani2007detecting,chan2012counting,chen2012feature, paragios2001mrf, marana1998efficacy,lempitsky2010learning,pham2015count,wang2016fast,xu2016crowd}.
These traditional methods' performance is limited by the weak feature representation power and the foreground/background extraction result.
To deal with the issues of traditional methods, deep learning methods are explored in the area.
\citet{zhang2019wide,zhang2022wide} proposed the first end-to-end DNNs-based framework for multi-view crowd counting and a large city-scene multi-view vision dataset CityStreet.
\citet{zhang2020_3d, zhang2022_3d} proposed to solve the problem in 3D space with the 3D feature fusion and the 3D density map supervision.
\citet{zhang2021cross} proposed a large synthetic multi-view dataset CVCS to handle the cross-view cross-scene setting,
and the method is applied to novel new scenes with domain transferring steps.
\citet{zheng2021learning} improved the late fusion model \cite{zhang2019wide} by introducing the correlation between each pair of views.
\citet{zhai2022co} proposed a graph-based multi-view learning model for multi-view counting.
Multi-view counting methods mainly focus on predicting crowd density maps on the ground and the people count but with relatively weak localization ability.


\textbf{Multi-camera tracking}.
Multi-camera tracking can track the objects under multi-cameras to deal with occlusions or lighting variations \cite{iguernaissi2019people}.
The existing methods can be categorized into centralized methods (overlapped) \cite{chavdarova2018wildtrack,fleuret2007multicamera,xu2016multi,you2020real}
and distributed methods (non-overlapped) \cite{patino2014multicamera,taj2011distributed,yang2022distributed}.
Here, we mainly review centralized methods with overlapping camera views.
Centralized methods consist of 3 steps: camera view people detection/feature extraction, data fusion and tracking.
\citet{you2020real} followed the steps and proposed a real-time 3D multi-camera tracking by fusing 2D people location predictions on the ground plane
and then tracking each person from the fused ground-plane maps.
\citet{nguyen2022lmgp} proposed to match the multi-camera trajectories by solving a global lifted multicut problem.

In summary, the model generalization ability has been explored in other multi-view vision tasks, such as using large synthetic datasets in training.
But in the area of multi-view people detection, the methods are only evaluated on the same single scenes due to
limited data, which reduces the model generalization potential under real-world application scenarios.
\emph{And no methods have tried estimating the view weights for fusion with the guidance of single-view ground-plane ground-truth,
requiring no extra labels.}

\section{Method}

In this section, we describe the proposed supervised view-wise contribution weighting multi-view detection method,
which consists of 4 stages (see Figure~\ref{fig:pipeline}):
\emph{Single-view feature extraction and projection, Projected single-view decoding, Supervised view-wise contribution weighted fusion, and Multi-view feature decoding}.
We first introduce the whole model's subnets and modules, where the details about the proposed supervised view-wise contribution weight module are presented.
Finally, we describe how we generalize the trained model to novel new scenes.

\subsection{Single-View Feature Extraction and Projection}
We choose ResNet \cite{he2016deep}/VGG \cite{simonyan2014very} as the feature extraction backbone net for the multi-view people detection model.
To handle the variable numbers of camera views in the training and testing scenes, the feature extraction subnet is shared across all input camera views.
After feature extraction, each view's features are projected to the scene ground plane for further processing via a projection layer with camera calibrations based on spatial transformation network \cite{jaderberg2015spatial}.
The projection layer implemented in our model could be used with variable camera parameters instead of a fixed set of ones to
handle camera view number change across different scenes.

\subsection{Projected Single-View Decoding}
\begin{figure}[t]
\begin{center}
   \includegraphics[width=\linewidth]{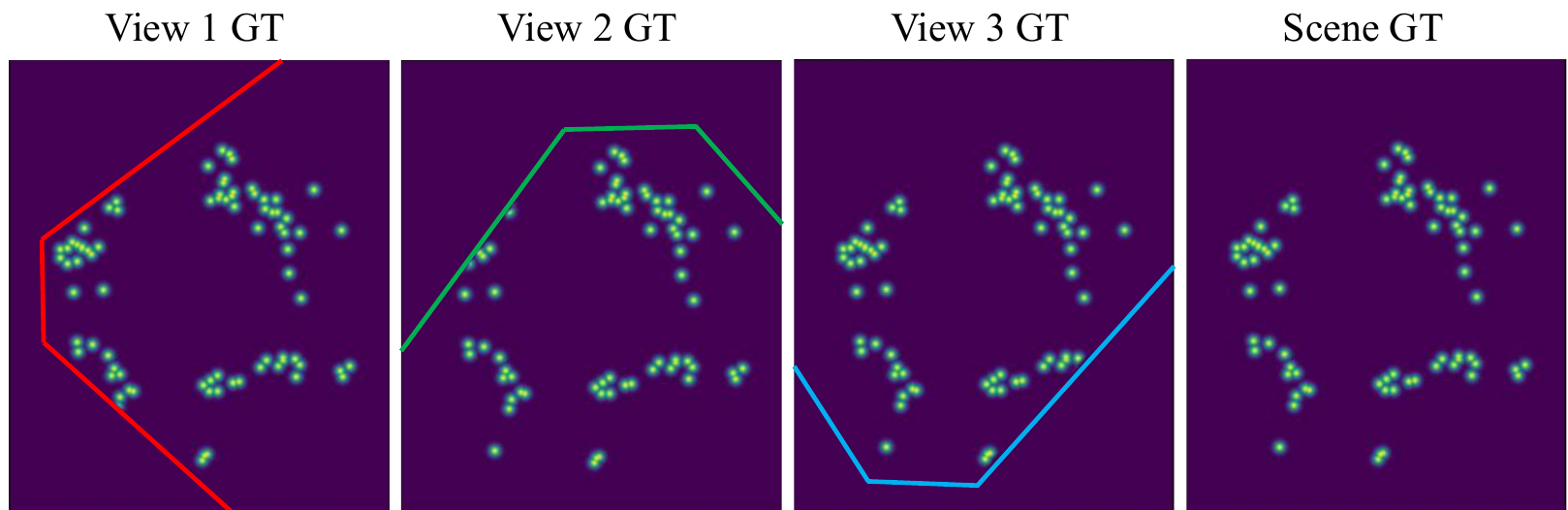}
\end{center}
   \caption{`View GT' is the ground-truth for each view in projected single-view decoding, which is the people occupancy map on the ground that
   can be seen by the corresponding view, and `Scene GT' stands for the ground-truth for the whole scene of CityStreet.
   The lines in the `View GT' indicate the field-of-view region of the camera view.
   }
\label{fig:view_scene_gt}
\end{figure}

We use a subnet to obtain each view's people location prediction on the ground plane based on the projected single-camera view features, which is shared
across all input camera views to handle the possible variable camera views.
The supervision for the decoding subnet training is the scene location map consisting of people that can be seen within the corresponding camera view.
Since the decoding result only contains people that can be seen in the field-of-view of each camera
(as shown in Figure~\ref{fig:view_scene_gt}), the prediction can be used as the confidence of the view on the corresponding regions in the final result.
So, we use the single-view ground-plane prediction results to fuse the multi-camera information in the next step.
Besides, the projected single-view decoding module also provides an extra constraint on the training of the model for the feature extraction module.
Thus, the feature extracted from the multi-view images should be effective in the single-view decoding after projection.

The projected single-view decoding loss $\ell_v$ can be calculated as follows. Denote $n$ as the camera view number, $i = 0,1,...,n-1$ stands for the index of
each view, and the prediction and ground truth for each view are $V_{i}$ and $V_{i}^{gt}$, respectively.
\begin{align}
\ell_v = \frac{1}{n}\sum_i{\lVert V_{i}\!-\!V_{i}^{gt}\rVert_2^2}= \frac{1}{n}\sum_i{\lVert V_{i}\!-\!V_{s}^{gt}\!\otimes \!M_i\rVert_2^2}.
\end{align}
$V_{i}^{gt}=V_{s}^{gt}\otimes M_i$ means each view's ground-truth in the projected single view is the
scene-level ground-truth $V_{s}^{gt}$ multiplied by the view's field-of-view mask $M_i$ on the ground.

\subsection{Supervised View-Wise Contribution Weighted Fusion}

We propose the supervised view-wise contribution weighted fusion approach for fusing multi-camera information.
First, each view's scene ground-plane prediction result $V_i$ is fed in the shared subnet $\calC$ to predict the weight map $\hat{W}_i$ for each camera view.
Then, the weight maps $\{\hat{W}_i\}$ for all views are normalized to make sure the sum of the weights for all camera views of one pixel on the
scene ground-plane map equals to 1, denoted as ${W}_i$.
In addition to that, the regions that cannot be seen by a camera view are assigned to $0$ weight under that view.
Especially in the normalization process, these regions' weights are not calculated in the final result.
Therefore, the view-wise field-of-view mask $M_i$ is multiplied with each camera view's initial weight map $\hat{W}_i$ before the normalization.
The process of the view-wise contribution weight maps can be calculated as follows.
\begin{equation}	
\begin{aligned}
  \hat{W}_i = \calC(V_i), W_i =\frac{{\hat{W}_i}\otimes M_i}{\sum_i{{\hat{W}_i}\otimes M_i + \sigma}} ,
\end{aligned}
\end{equation}
where $\sigma$ is a small value to avoid the zero denominator issue when a region pixel cannot be seen by any input views.

After that, each camera view's projected features $F_i$ are multiplied with the view-wise contribution weight maps ${W_i}$,
and summed together to obtain the scene-level feature representation $F = \sum_i{{F_i}\otimes {W_i}}$.
\emph{To the best of our knowledge, this is the first work that uses the supervised view-wise contribution on the scene ground-plane map as a weighting method for fusing multi-camera view information in the field, which provides more guidance of the people contained in each view. Compared to other weighted methods, SHOT \cite{song2021stacked} or MVDeTr \cite{hou2021multiview}, the proposed method is more stable on different datasets (see experiment section for more details).}

\subsection{Multi-View Feature Decoding}
After obtaining the fused feature representation $F$ for multi-cameras, $F$ is fed in a decoder for predicting the scene-level prediction $V_s$
of the people occupancy map on the ground.
Note this decoder is different from the one used for projected single-view decoding because they are targeting different functions,
one for decoding each camera view's features, and the other one for the whole scene's feature representation.
The $\mse$ loss is also used in the multi-view feature decoding, denoted as $\ell_s=\mse(V_{s}, V_{s}^{gt})$.
And together with the projected single-view decoding loss $\ell_v$, the model's loss $\ell$ can be summarised as $ \ell = \ell_s + \lambda \ell_v$,
where $\lambda$ is used to adjust the two decoding losses' importance in the training.

\begin{figure}[t]
\begin{center}
   \includegraphics[width=\linewidth]{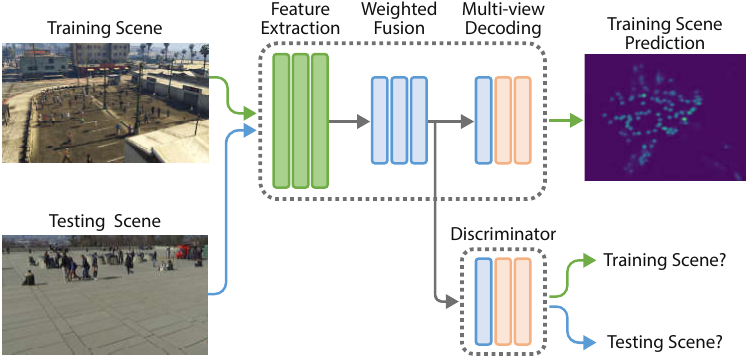}
\end{center}
   \caption{The domain adaptation approach used in our method for generalizing to novel new scenes.}
\label{fig:domain_adaptation}
\end{figure}

\subsection{Generalization to New Scenes}
Our proposed supervised view-wise contribution weighting method is trained on a large synthetic multi-view people dataset CVCS \cite{zhang2021cross}, which can be applied to new scenes with promising results by slightly finetuning the model. To further reduce the large domain gap between the training scenes and testing new scenes, we also use a domain adaptation method to improve the performance (see Figure~\ref{fig:domain_adaptation}) after finetuning the trained model on the new scenes with limited labeled data.
In particular, we add a discriminator in the trained model to reduce the gap between the training scene features and testing scene features. In the finetuning stage, we first trained the model by using 5\% of the new scene training set images, and then both the training synthetic images and the testing new scene images are fed into the proposed model. Finally, both kinds of features are classified by the discriminator. The loss in the finetuning includes the new scene multi-view detection loss, the synthetic multi-view detection loss, and the discriminator classification loss. In experiments, the model's cross-scene multi-view detection performance is promising compared to the previous methods trained on the same single scenes, which can extend the multi-view people detection to more general application scenarios.

\begin{table}[t]
  \centering
  \small

  \begin{tabular}{@{}l@{\hspace{0.08cm}}c@{\hspace{0.08cm}}c@{\hspace{0.08cm}}c@{\hspace{0.08cm}}c@{\hspace{0.08cm}}c@{\hspace{0.08cm}}c@{}}
  \hline
   Dataset     & Frames   & Scene  & Resolution     & Counts & Views  & Area  \\
  \hline
  CVCS &    200k/80k  & 23/8   & $1920\!\times\!1080$           & 90-180 & 60-120 & 90$\times$80  \\
  CityStreet & 300/200  & 1    & $2704\!\times\!1520$           & 70-150 & 3      &  58$\times$72   \\ 

  Wildtrack    & 360/40 & 1  & $1920\!\times\!1080$   & 20  & 7      & 12$\times$36 \\ 
  MultiviewX    & 360/40  & 1 &$1920\!\times\!1080$     & 40  & 6      & 16$\times$25 \\ 
  \hline
  \end{tabular}
  \caption{Comparison of multi-view people datasets. `/' stands for the training and testing statistics.}
    \label{tab:compare_dataset}
\end{table}

\section{Experiments and Results}
In this section, we first introduce the datasets used in the experiments and then present the experiment settings, including the comparison methods,
the implementation details, and evaluation metrics. Finally, we show and compare the experiment results, including the multi-view people detection
performance on various datasets and the ablation study on the proposed view-wise contribution weighting module.

\subsection{Datasets}
We introduce 4 datasets used in the multi-view people detection, including CVCS \cite{zhang2021cross}, CityStreet \cite{zhang2019wide},
 Wildtrack \cite{chavdarova2018wildtrack} and MultiviewX \cite{hou2020multiview}, among which the latter 2 datasets are relatively smaller in the scene size (see dataset comparison in Table \ref{tab:compare_dataset}).
\textbf{CVCS} is a synthetic multi-view people dataset, containing 31 scenes, where 23 are for training and the rest 8 for testing. The scene size varies from about $10m*20m$ to $90m*80m$.
Each scene contains 100 multi-view frames.
The ground plane map resolution is $900\!\times\!800$, where each grid stands for 0.1 meter in the real world.
In the training, 5 views are randomly selected for 5 times in each iteration per frame of each scene, and the same view number is randomly selected for 21 times in evaluation.
\textbf{CityStreet} is a real-world city scene dataset collected around the intersection of a crowded street.
The scene size of the dataset is around $58m\!\times\!72m$.
The ground plane map resolution is $320\!\times\!384$.
\textbf{Wildtrack} is a real-world dataset recorded on the square of a university campus.
The ground plane map resolution $120\!\times\!360$, where each grid stands for 0.1m in the real world.
\textbf{MultiviewX} is a synthetic dataset for multi-view people detection.
The ground plane map resolution is $250\!\times\!160$, where each grid also stands for 0.1m in the real world.

Compared to Wildtrack and MultiviewX, CVCS and CityStreet contain more scenes, more camera views and more images, which
are more suitable for validating multi-view people detection tasks in more practical environments.
Thus, unlike other methods, we mainly evaluate on larger datasets CVCS and CityStreet.

\begin{figure}[t]
\begin{center}
   \includegraphics[width=\linewidth]{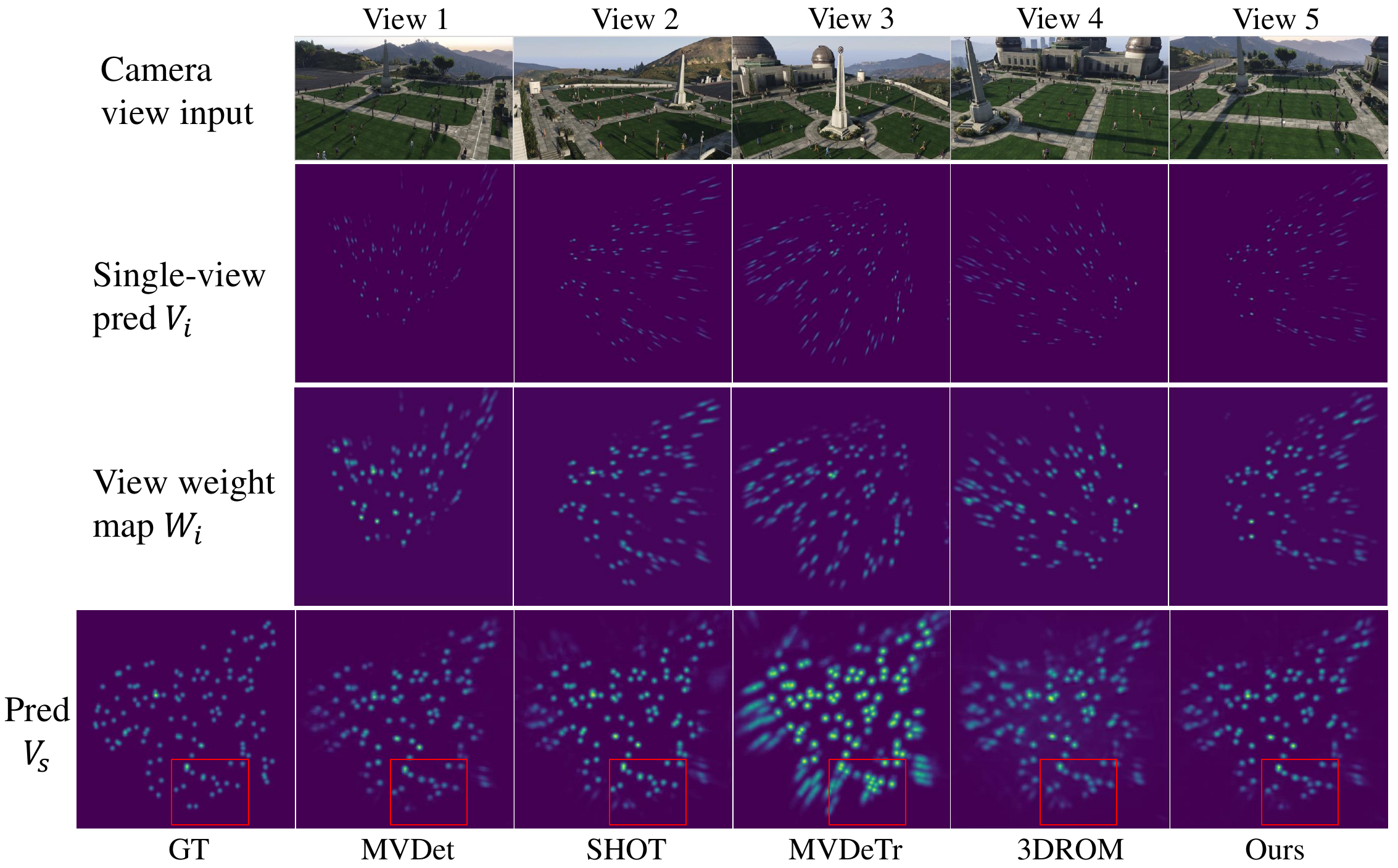}
\end{center}
   \caption{The result visualization of the method: camera view input, single-view prediction, view weight map and the corresponding ground-truth and prediction results.}
\label{fig:result_vis}
\end{figure}

\begin{table*}[t]
\small
\centering
\begin{tabular}{l@{\hspace{0.12cm}}|c@{\hspace{0.12cm}}c@{\hspace{0.12cm}}c@{\hspace{0.12cm}}c@{\hspace{0.12cm}}c@{\hspace{0.12cm}}c@{\hspace{0.12cm}}|
c@{\hspace{0.12cm}}c@{\hspace{0.12cm}}c@{\hspace{0.12cm}}c@{\hspace{0.12cm}}c@{\hspace{0.12cm}}c@{\hspace{0.12cm}}|c@{\hspace{0.12cm}}}
\hline
    Dataset &  \multicolumn{6}{c|}{CVCS}  &  \multicolumn{6}{c|}{CityStreet}  &  \multicolumn{1}{c}{} \\
    Method                                   & MODA    & MODP        & Precision        & Recall     & F1\_score  & Rank
                                             & MODA    & MODP        & Precision        & Recall     & F1\_score  & Rank   & Avg. Rank\\
\hline
    MVDet          & 36.6    & 71.0     & 79.4     & 49.4      & 60.9  & 4  & 44.6    & 65.7     & 79.8     & 59.8      & 68.4         & 5  & 4.5\\  
    SHOT            & 45.0  & 77.4     & 83.6    & 55.9      & \underline{67.0} & \underline{2}  & 53.5    & 72.4     & 91.0     & 59.4   & 71.8   &  4 &  3 \\ 
    MVDeTr       & 39.8    & 84.1     & 95.3     & 44.9      & 61.0  & 3  & 58.3    & 74.1     & 92.8     & 63.2      & 75.2           &  3 &  3\\ 
    3DROM        & 33.9    & 73.9    & 79.5     & 42.2      &55.1  & 5  & 60.0    & 70.1     & 82.5     & 76.2      & \textbf{79.2}  &  \textbf{1} &  3\\ 
\hline
\hline
    Ours        & 46.2	& 78.4     & 81.2     & 59.1      & \textbf{68.4} & \textbf{1}  & 55.0	& 70.0     & 81.4     & 71.2     & \underline{76.0}  & \underline{2} &  \textbf{1.5}\\
\hline
\end{tabular}

\caption{Comparison of the multi-view people detection performance on the larger datasets CVCS and CityStreet using 5 metrics. The distance threshold is $1$m on CVCS (5 pixels on the ground plane map), and $2$m on CityStreet (20 pixels on the ground plane map).  See results with other distance thresholds in the supplemental.
Overall, all previous methods do not perform well on the 2 large datasets compared to Wildtrack and MultiviewX (see in Table \ref{table:Wildtrack_results}).
The proposed method ranks the best among all methods according to the average rank on the 2 datasets.}
\label{table:CVCS_results}
\end{table*}

\begin{table}
\small


\centering
\begin{tabular}{l|c@{\hspace{0.15cm}}c@{\hspace{0.15cm}}c@{\hspace{0.15cm}}c@{\hspace{0.15cm}}c@{\hspace{0.15cm}}c@{\hspace{0cm}}}
\hline
Backbone    & Method         & MODA    & MODP        & P.        & R.     & F1.     \\
\hline
\multirow{2}{*}{ResNet}
    &With            & \textbf{46.2}   & \textbf{78.4}    & \textbf{81.2}        & \textbf{59.1}  & \textbf{68.4} \\
    &Without          & 36.6   & 71.0   & 79.4       & 49.4  & 60.9  \\
\hline
\multirow{2}{*}{VGG}
    &With               & \textbf{39.9}    & 71.9     & 85.7        & \textbf{47.9}  & \textbf{61.5}  \\
    &Without          & 38.1    & \textbf{77.1}     & \textbf{86.3}        & 45.3  & 59.4  \\

\hline
\end{tabular}
\caption{The ablation study on whether the proposed supervised view-wise contribution weighted fusion is used or not (with/without) on CVCS dataset.}
\label{table:module_backbone_ablation}
\end{table}

\subsection{Experiment Settings}
\textbf{Comparison methods.}
We compare the proposed view-wise contribution weighting method with several state-of-the-art multi-view people detection methods:
MVDet (ECCV 2020) \cite{hou2020multiview}, SHOT (ICCV 2021) \cite{song2021stacked}, MVDeTr (ACM MM 2021) \cite{hou2021multiview},
and 3DROM (ECCV 2022) \cite{qiu20223d}.
We run these four latest multi-view people detection methods on large multi-view people datasets CVCS and CityStreet,
using the code implemented by the corresponding paper authors.
We also compare with other methods, such as RCNN \cite{xu2016multi}, POM-CNN \cite{fleuret2007multicamera}, DeepMCD \cite{chavdarova2017deep},
DeepOcc. \cite{baque2017deep}, and Volumetric \cite{iskakov2019learnable}, on Wildtrack and MultiviewX.


\textbf{Implementation details.}
The proposed model is based on ResNet/VGG backbone. For model setting, the layer setting of feature extraction and decoders for projected single-view
decoding and multi-view decoding can be found in the supplemental.
For the view-wise contribution weighted fusion, the single-view predictions are fed into a 4-layer subnet:
$[3\!\times\!3\!\times\!1\!\times\!256, 3\!\times\!3\!\times\!256\!\times\!256, 3\!\times\!3\!\times\!256\!\times\!128, 3\!\times\!3\!\times\!128\!\times\!1]$.
The map classification threshold is 0.4 for all datasets, and the distance threshold is 1m (5 pixels) on CVCS, 2m (20 pixels) on CityStreet, and 0.5m (5 pixels) on MultiviewX and Wildtrack.
As to the  model training, a 3-stage training is used:
First, the 2D counting task is trained as the pretraining for the feature extraction subnet; 
Then, the projected single-view decoding subnet is trained after loading the pre-trained feature extraction subnet;
Finally, the projected single-view decoding subnet and the multi-view decoding subnet are trained together, where the loss term weight $\lambda=1$.
We follow other training settings as in MVDet.

\textbf{Evaluation metrics.}
We use 5 metrics to evaluate and compare the multi-view people detection methods:
Multiple Object Detection Accuracy (MODA),
Multiple Object Detection Precision (MODP),
Precision, Recall and F1\_score.
We calculate true positive (TP), false positive (FP), and false negative (FN) first to
compute the metrics. 
$MODA = 1 - (FP+FN)/(TP+FN)$, shows the detection accuracy.
$MODP = (\sum(1-d[d<t]/t))/TP$, shows the precision of detection, where $d$ is the distance from a detected person point to
its ground truth and $t$ is the distance threshold.
$Precion = TP/(FP+TP)$, $Recall = TP/(TP+FN)$, and $F1\_score = 2Precion*Recall/(Precion+Recall)$,
where $F1\_score$ is a balance of $Precion$ and $Recall$ for detection performance evaluation.
Additionally, the \textbf{Rank} and the average rank (\textbf{Avg. Rank}) of each method's performance on CVCS and CityStreet are also presented to compare different methods' overall performance.

\subsection{Experiment Results}

We show the performance on CVCS and CityStreet in Table \ref{table:CVCS_results}. 
Overall, compared to results on Wildtrack and MultiviewX (see in Table \ref{table:Wildtrack_results}),
the performance on large scenes, CVCS and CityStreet, is much lower.
On \textbf{CVCS}, compared with all other methods, our proposed method achieves the best performance.
The proposed method shares the same backbone model with the MVDet method \cite{hou2020multiview},
but our overall performance is better than MVDet, which shows the effectiveness of the proposed method.
SHOT uses an extra multi-height projection and works well when calibration errors of the dataset are relatively small as
in CVCS, and it performs much worse on CityStreet due to CityStreet having larger calibration errors, causing extra difficulties for the multi-height fusion.
3DROM is better than our method on CityStreet because it is a data augmentation method that deals with the data lacking issue better. But 3DROM works badly on CVCS because CVCS is already a very large dataset containing various camera and scene variations.
On \textbf{CityStreet}, the proposed method (using VGG as the backbone) also achieves the second-best performance according to F1\_score metric, which is better than SHOT \cite{song2021stacked}, MVDeTr \cite{hou2021multiview} and MVDet \cite{hou2020multiview}.
In addition, MVDeTr utilizes deformable transformer modules, which are relatively easy to learn on small datasets. However, on large datasets like CVCS with a high number of camera views that keep changing during training, it has difficulty stabilizing the weight learning process, limiting its detection performance

Overall, the proposed supervised view-wise contribution weighting method achieves the best average rank (Avg. Rank) among all methods.
The reason is the view-wise ground-plane supervision provides more clues for the people locations of each view, and thus the multi-view fusion performance is more stable and better than other methods. We also show the visualization result on CVCS dataset in Figure~\ref{fig:result_vis}, where the first 3 rows are the multi-view inputs, the proposed method' single-view predictions, and the view weight maps, indicating accurate people ground locations.

\begin{table}[t]
\small
\centering
\begin{tabular}{l|c@{\hspace{0.15cm}}c@{\hspace{0.15cm}}c@{\hspace{0.15cm}}c@{\hspace{0.15cm}}c@{\hspace{0.15cm}}c@{\hspace{0.10cm}}}
\hline
Dataset    & Method         & MODA    & MODP        & P.        & R.     & F1.     \\
\hline
\multirow{2}{*}{CVCS}

    & Supervised          & \textbf{46.2}    & \textbf{78.4}    & 81.2        & \textbf{59.1}  & \textbf{68.4} \\
    & Unsupervised         & 45.8	& 73.6     & \textbf{86.7}     & 54.1      & 66.6        \\
\hline
\multirow{2}{*}{CityStreet}
    & Supervised          & \textbf{55.0}	& \textbf{70.0}     & \textbf{81.4}     & \textbf{71.2}      & \textbf{76.0}  \\
    & Unsupervised         & 49.5	& 67.1     & 78.3     & 68.5      & 73.1        \\

\hline
\end{tabular}
\caption{The ablation study on whether the view-wise contribution weighted fusion is supervised or unsupervised. 
}
\label{table:supervision_ablation}
\end{table}

\begin{table}[t]
\small
\centering
\begin{tabular}{c|c@{\hspace{0.15cm}}c@{\hspace{0.15cm}}c@{\hspace{0.15cm}}c@{\hspace{0.15cm}}c@{\hspace{0.10cm}}}
\hline
    Num.         & MODA    & MODP        & Precision        & Recall     & F1\_score     \\
\hline
    3            & 37.1    & 73.4       & 70.7            & 62.1        & 66.1     \\
    5            & 46.2    & 78.4       &  81.2           & 59.1      & 68.4    \\
    7            & 50.5    & 76.6        & 90.1            & 56.8        & 69.7     \\
    9            & 50.3    & 78.3        & 92.5            & 54.7        & 68.8     \\
\hline
\end{tabular}

\caption{The ablation study on the variable testing camera number (3, 5, 7, 9) of the proposed method on the CVCS dataset, which is trained on 5 camera views.}
\label{table:camNum_ablation}
\end{table}

\begin{table*}[t]
\small

\centering
\begin{tabular}{l@{\hspace{0.12cm}}|c@{\hspace{0.12cm}}c@{\hspace{0.12cm}}c@{\hspace{0.12cm}}c@{\hspace{0.12cm}}c@{\hspace{0.12cm}}|c@{\hspace{0.12cm}}
c@{\hspace{0.12cm}}c@{\hspace{0.12cm}}c@{\hspace{0.12cm}}c@{\hspace{0.12cm}}}
\hline
    Dataset &  \multicolumn{5}{c|}{Wildtrack}  &  \multicolumn{5}{c}{MultiviewX}  \\
    Method         & MODA    & MODP        & Precision        & Recall     & F1\_score
                     & MODA    & MODP        & Precision        & Recall     & F1\_score \\
\hline
    RCNN \cite{xu2016multi}                    & 11.3    & 18.4        & 68         &  43  & 52.7
                                                   & 18.7    & 46.4        & 63.5         &  43.9  & 51.9 \\
    POM-CNN \cite{fleuret2007multicamera}      &  23.2  &    30.5      & 75     & 55  & 63.5
                                                   &  -  &    -      & -     & -  & -       \\
    DeepMCD  \cite{chavdarova2017deep}          & 67.8   & 64.2         & 85         & 82  & 83.5
                                                   & 70.0   & 73.0         & 85.7         & 83.3  & 84.5\\
    DeepOcc. \cite{baque2017deep}              & 74.1    & 53.8        & 95     & 80  & 86.9
                                                   & 75.2    & 54.7        & 97.8     & 80.2  & 88.1\\
    Volumetric \cite{iskakov2019learnable}     & 88.6    & 73.8        & 95.3     & 93.2  & 94.2
                                                   & 84.2    & 80.3        & 97.5     & 86.4  & 91.6\\
\hline
    MVDet   \cite{hou2020multiview}              & 88.2    & 75.7        & 94.7        & 93.6  & 94.1
                                                   & 83.9    & 79.6        & 96.8        & 86.7  & 91.5\\
    SHOT    \cite{song2021stacked}                & 90.2    & 76.5        & 96.1        & 94.0  & 95.0
                                                   & 88.3    & 82.0        & 96.6        & 91.5  & 94.0 \\
    MVDeTr  \cite{hou2021multiview}            & 91.5    & 82.1        & 97.4        & 94.0  & 95.7
                                                    & 93.7    & 91.3        & 99.5        & 94.2  & 97.8 \\
    3DROM   \cite{qiu20223d}                     & 93.5    & 75.9        & 97.2        & 96.2  & 96.7
                                                     & 95.0    & 84.9        & 99.0        & 96.1  & 97.5 \\
\hline
\hline
    Ours (ft)                                  & 73.9	& 72.4          & 86.8       & 87.2   & 87.0
                                              & 81.1	& 77.2         & 95.0        & 85.6 	& 90.1 \\
    Ours (ft+da)                          & \textbf{78.9}	& 73.6          & 88.7       & 90.4   & \textbf{89.5}
                                        & \textbf{83.8}	& 76.5         & 97.1        & 86.4   & \textbf{91.4} \\
\hline
\end{tabular}

\caption{Comparison of the multi-view people detection performance on Wildtrack and MultiviewX using 5 metrics.
All comparison methods train and test on Wildtrack or MultiviewX (single scene), while ours are trained on CVCS and finetuned on Wildtrack or MultiviewX with limited labeled data (`Ours (ft)') or with the domain adaptation technique (`Ours (ft+da)').
}
\label{table:Wildtrack_results}
\end{table*}




\subsection{Ablation Study}

\textbf{With/without the supervised view-wise contribution weighted fusion.}
The first ablation study is on the effectiveness of the proposed supervised view-wise contribution weighted fusion.
As shown in Table \ref{table:module_backbone_ablation}, no matter which backbone is used, the model with the supervised view-wise contribution weighted fusion
achieves better overall performance than the model without using it, which demonstrates the proposed approach's effectiveness.

\textbf{Supervised/unsupervised view-wise contribution weighted fusion.}
The second ablation study is on whether the view-wise contribution weighted fusion module is supervised or unsupervised. As shown in Table \ref{table:supervision_ablation}, on both CVCS and CityStreet datasets, the supervised view-wise contribution weighted fusion achieves better results than the unsupervised one. The reason is the supervised one provides extra guidance for each view and it's beneficial for better multi-view fusion results. Note that the supervision for each view is obtained from the scene-level ground-truth, and no extra labeling efforts are required.



\textbf{Variable camera number.}
The fourth ablation study is on the variable camera number in the testing stage.
To generalize the model to novel new scenes requires that the model can be applied to variable camera view number inputs,
because real testing scenes may contain different numbers of camera views.
The proposed method is trained on 5-camera-view inputs on CVCS dataset \cite{zhang2021cross} while tested on variable camera view number inputs,
namely 3, 5, 7, and 9. Note that the ground-truth for each testing setting is the people captured by the variable camera views.
As shown in Table \ref{table:camNum_ablation}, while the camera view number is increased from 3 to 9, the MODA, MODP, and Precision metrics are also generally increasing,
while the Recall metric is decreasing. The reason is, that when the camera view number is increased, the model can detect more TP cases with higher accuracy.
But increasing the camera view number also means more people need to be detected (ground-truth people number increases), which causes more FN cases, too,
and thus the Recall metric decreases. But overall, the F1\_score is stable (decreases a little), which shows the model is relatively stable across camera view number changes.

\textbf{Generalization to new scenes.}
We show the cross-scene performance of the proposed method on Wildtrack and MultiviewX in Table \ref{table:Wildtrack_results},
which is trained on the large dataset CVCS.
We first finetune the trained model on Wildtrack and MultiviewX by using 5\% of the training set images (`Ours (ft)') from the new scenes, then use a domain adaptation approach \cite{tzeng2017adversarial} to reduce the domain gap between source and target scenes and further improve the performance (`Ours (ft+da)').
From  Table \ref{table:Wildtrack_results}, `Ours (ft)' already outperforms 4 comparison methods which use 100\% training set data and tested on the same single scene:
RCNN \cite{xu2016multi}, POM-CNN \cite{fleuret2007multicamera}, DeepMCD  \cite{chavdarova2017deep}, DeepOcc. \cite{baque2017deep}.
With the domain adaptation approach, the target and source domain gap is reduced and the cross-scene performance is further improved on both datasets. On MultiviewX (with a larger crowd number than Wildtrack), `Ours (ft+da)' achieves close performance to the state-of-the-art methods MVDet \cite{hou2020multiview} and Volumetric \cite{iskakov2019learnable}.
Compared to the rest methods, the proposed method's cross-scene performance with unsupervised domain adaptation (`Ours (ft+da)') is relatively worse, but considering that our method uses only 5\% of the target scene labels and achieves very close performance to other state-of-the-art methods using 100\% training set data and testing on the same single scenario, the proposed method's result is still promising.

\section{Discussion and Conclusion}

In this paper, we present a novel supervised view-wise contribution weighting approach for multi-view people detection in large scenes.
We evaluate its performance on large multi-view datasets, which is a departure from the typical approach of using small single-scene datasets.
We have demonstrated that our proposed method performs better on larger and more complicated scenes, and achieves promising cross-scene multi-view people detection performance compared with existing state-of-the-art techniques trained on single scenes.
To our knowledge, this is the first study on the large-scene multi-view people detection task.
Our proposed method extends the applicability of multi-view people detection to more practical scenarios, making it a valuable tool for
various applications in the fields of computer vision, surveillance, and security.
\textbf{Limitations}: The adopted domain transferring method is simple and limited by the image style transferring a lot. A stronger domain-transferring module could be our future work.

\section{Ethical Statement}
We use four datasets CVCS, CityStreet, MultiviewX, and Wildtrack in the experiments, among which CVCS and MultiviewX are synthetic datasets and the rest 2 are public real-scene datasets.

\section{Acknowledgements}
This work was supported in parts by NSFC (62202312, 62161146005, U21B2023, U2001206), DEGP Innovation Team (2022KCXTD025), CityU Strategic Research Grant (7005665), and Shenzhen Science and Technology Program (KQTD20210811090044003, RCJC20200714114435012, JCYJ20210324120213036).

\bibliography{aaai24}

\begin{thebibliography}{47}
\providecommand{\natexlab}[1]{#1}

\bibitem[{Baqu{\'e}, Fleuret, and Fua(2017)}]{baque2017deep}
Baqu{\'e}, P.; Fleuret, F.; and Fua, P. 2017.
\newblock Deep occlusion reasoning for multi-camera multi-target detection.
\newblock In \emph{Proceedings of the IEEE International Conference on Computer
  Vision}, 271--279.

\bibitem[{Chan and Vasconcelos(2012)}]{chan2012counting}
Chan, A.~B.; and Vasconcelos, N. 2012.
\newblock Counting people with low-level features and Bayesian regression.
\newblock \emph{IEEE Transactions on Image Processing}, 21(4): 2160--2177.

\bibitem[{Chavdarova et~al.(2018)Chavdarova, Baqu{\'e}, Bouquet, Maksai, Jose,
  Bagautdinov, Lettry, Fua, Van~Gool, and Fleuret}]{chavdarova2018wildtrack}
Chavdarova, T.; Baqu{\'e}, P.; Bouquet, S.; Maksai, A.; Jose, C.; Bagautdinov,
  T.; Lettry, L.; Fua, P.; Van~Gool, L.; and Fleuret, F. 2018.
\newblock WILDTRACK: A Multi-camera HD Dataset for Dense Unscripted Pedestrian
  Detection.
\newblock In \emph{Proceedings of the IEEE Conference on Computer Vision and
  Pattern Recognition}, 5030--5039.

\bibitem[{Chavdarova and Fleuret(2017)}]{chavdarova2017deep}
Chavdarova, T.; and Fleuret, F. 2017.
\newblock Deep multi-camera people detection.
\newblock In \emph{2017 16th IEEE international conference on machine learning
  and applications (ICMLA)}, 848--853. IEEE.

\bibitem[{Chen et~al.(2012)Chen, Chen, Gong, and Xiang}]{chen2012feature}
Chen, K.; Chen, L.~C.; Gong, S.; and Xiang, T. 2012.
\newblock Feature mining for localised crowd counting.
\newblock In \emph{BMVC}.

\bibitem[{Cheng et~al.(2022)Cheng, Dai, Li, Song, Wu, and
  Hauptmann}]{cheng2022rethinking}
Cheng, Z.-Q.; Dai, Q.; Li, H.; Song, J.; Wu, X.; and Hauptmann, A.~G. 2022.
\newblock Rethinking spatial invariance of convolutional networks for object
  counting.
\newblock In \emph{Proceedings of the IEEE/CVF Conference on Computer Vision
  and Pattern Recognition}, 19638--19648.

\bibitem[{Cheng et~al.(2019{\natexlab{a}})Cheng, Li, Dai, Wu, and
  Hauptmann}]{cheng2019learning}
Cheng, Z.-Q.; Li, J.-X.; Dai, Q.; Wu, X.; and Hauptmann, A.~G.
  2019{\natexlab{a}}.
\newblock Learning spatial awareness to improve crowd counting.
\newblock In \emph{Proceedings of the IEEE/CVF international conference on
  computer vision}, 6152--6161.

\bibitem[{Cheng et~al.(2019{\natexlab{b}})Cheng, Li, Dai, Wu, He, and
  Hauptmann}]{cheng2019improving}
Cheng, Z.-Q.; Li, J.-X.; Dai, Q.; Wu, X.; He, J.-Y.; and Hauptmann, A.~G.
  2019{\natexlab{b}}.
\newblock Improving the Learning of Multi-column Convolutional Neural Network
  for Crowd Counting.
\newblock In \emph{Proceedings of the 27th ACM International Conference on
  Multimedia}, 1897--1906.

\bibitem[{Fleuret et~al.(2007)Fleuret, Berclaz, Lengagne, and
  Fua}]{fleuret2007multicamera}
Fleuret, F.; Berclaz, J.; Lengagne, R.; and Fua, P. 2007.
\newblock Multicamera people tracking with a probabilistic occupancy map.
\newblock \emph{IEEE transactions on pattern analysis and machine
  intelligence}, 30(2): 267--282.

\bibitem[{Gall et~al.(2011)Gall, Yao, Razavi, Van~Gool, and
  Lempitsky}]{gall2011hough}
Gall, J.; Yao, A.; Razavi, N.; Van~Gool, L.; and Lempitsky, V. 2011.
\newblock Hough forests for object detection, tracking, and action recognition.
\newblock \emph{IEEE transactions on pattern analysis and machine
  intelligence}, 33(11): 2188--2202.

\bibitem[{He et~al.(2016)He, Zhang, Ren, and Sun}]{he2016deep}
He, K.; Zhang, X.; Ren, S.; and Sun, J. 2016.
\newblock Deep residual learning for image recognition.
\newblock In \emph{Proceedings of the IEEE conference on computer vision and
  pattern recognition}, 770--778.

\bibitem[{Hou and Zheng(2021)}]{hou2021multiview}
Hou, Y.; and Zheng, L. 2021.
\newblock Multiview detection with shadow transformer (and view-coherent data
  augmentation).
\newblock In \emph{Proceedings of the 29th ACM International Conference on
  Multimedia}, 1673--1682.

\bibitem[{Hou, Zheng, and Gould(2020)}]{hou2020multiview}
Hou, Y.; Zheng, L.; and Gould, S. 2020.
\newblock Multiview detection with feature perspective transformation.
\newblock In \emph{Computer Vision--ECCV 2020: 16th European Conference,
  Glasgow, UK, August 23--28, 2020, Proceedings, Part VII 16}, 1--18. Springer.

\bibitem[{Huang et~al.(2020)Huang, Li, Cheng, Zhang, and
  Hauptmann}]{huang2020stacked}
Huang, S.; Li, X.; Cheng, Z.-Q.; Zhang, Z.; and Hauptmann, A. 2020.
\newblock Stacked pooling for boosting scale invariance of crowd counting.
\newblock In \emph{ICASSP 2020-2020 IEEE International Conference on Acoustics,
  Speech and Signal Processing (ICASSP)}, 2578--2582. IEEE.

\bibitem[{Iguernaissi et~al.(2019)Iguernaissi, Merad, Aziz, and
  Drap}]{iguernaissi2019people}
Iguernaissi, R.; Merad, D.; Aziz, K.; and Drap, P. 2019.
\newblock People tracking in multi-camera systems: a review.
\newblock \emph{Multimedia Tools and Applications}, 78: 10773--10793.

\bibitem[{Iskakov et~al.(2019)Iskakov, Burkov, Lempitsky, and
  Malkov}]{iskakov2019learnable}
Iskakov, K.; Burkov, E.; Lempitsky, V.; and Malkov, Y. 2019.
\newblock Learnable Triangulation of Human Pose.
\newblock In \emph{ICCV}.

\bibitem[{Jaderberg et~al.(2015)Jaderberg, Simonyan, Zisserman
  et~al.}]{jaderberg2015spatial}
Jaderberg, M.; Simonyan, K.; Zisserman, A.; et~al. 2015.
\newblock Spatial transformer networks.
\newblock In \emph{Advances in neural information processing systems},
  2017--2025.

\bibitem[{Joachims(1998)}]{joachims1998text}
Joachims, T. 1998.
\newblock Text categorization with support vector machines: Learning with many
  relevant features.
\newblock In \emph{European conference on machine learning}, 137--142.
  Springer.

\bibitem[{Lempitsky and Zisserman(2010)}]{lempitsky2010learning}
Lempitsky, V.; and Zisserman, A. 2010.
\newblock Learning to count objects in images.
\newblock In \emph{Advances in Neural Information Processing Systems},
  1324--1332.

\bibitem[{Marana et~al.(1998)Marana, Costa, Lotufo, and
  Velastin}]{marana1998efficacy}
Marana, A.; Costa, L. d.~F.; Lotufo, R.; and Velastin, S. 1998.
\newblock On the efficacy of texture analysis for crowd monitoring.
\newblock In \emph{International Symposium on Computer Graphics, Image
  Processing, and Vision}, 354--361. IEEE.

\bibitem[{Nguyen et~al.(2022)Nguyen, Henschel, Rosenhahn, Sonntag, and
  Swoboda}]{nguyen2022lmgp}
Nguyen, D.~M.; Henschel, R.; Rosenhahn, B.; Sonntag, D.; and Swoboda, P. 2022.
\newblock LMGP: Lifted Multicut Meets Geometry Projections for Multi-Camera
  Multi-Object Tracking.
\newblock In \emph{Proceedings of the IEEE/CVF Conference on Computer Vision
  and Pattern Recognition}, 8866--8875.

\bibitem[{Paragios and Ramesh(2001)}]{paragios2001mrf}
Paragios, N.; and Ramesh, V. 2001.
\newblock A MRF-based approach for real-time subway monitoring.
\newblock In \emph{Computer Vision and Pattern Recognition}, volume~1. IEEE.

\bibitem[{Patino and Ferryman(2014)}]{patino2014multicamera}
Patino, L.; and Ferryman, J. 2014.
\newblock Multicamera trajectory analysis for semantic behaviour
  characterisation.
\newblock In \emph{2014 11th IEEE International Conference on Advanced Video
  and Signal Based Surveillance (AVSS)}, 369--374. IEEE.

\bibitem[{Pham et~al.(2015)Pham, Kozakaya, Yamaguchi, and
  Okada}]{pham2015count}
Pham, V.-Q.; Kozakaya, T.; Yamaguchi, O.; and Okada, R. 2015.
\newblock Count forest: Co-voting uncertain number of targets using random
  forest for crowd density estimation.
\newblock In \emph{Proceedings of the IEEE International Conference on Computer
  Vision}, 3253--3261.

\bibitem[{Qiu et~al.(2022)Qiu, Xu, Yan, Smith, and Yang}]{qiu20223d}
Qiu, R.; Xu, M.; Yan, Y.; Smith, J.~S.; and Yang, X. 2022.
\newblock 3D Random Occlusion and Multi-layer Projection for Deep Multi-camera
  Pedestrian Localization.
\newblock In \emph{Computer Vision--ECCV 2022: 17th European Conference, Tel
  Aviv, Israel, October 23--27, 2022, Proceedings, Part X}, 695--710. Springer.

\bibitem[{Sabzmeydani and Mori(2007)}]{sabzmeydani2007detecting}
Sabzmeydani, P.; and Mori, G. 2007.
\newblock Detecting pedestrians by learning shapelet features.
\newblock In \emph{IEEE Conference on Computer Vision and Pattern Recognition},
  1--8. IEEE.

\bibitem[{Simonyan and Zisserman(2014)}]{simonyan2014very}
Simonyan, K.; and Zisserman, A. 2014.
\newblock Very deep convolutional networks for large-scale image recognition.
\newblock \emph{arXiv preprint arXiv:1409.1556}.

\bibitem[{Song et~al.(2021)Song, Wu, Yang, Zhang, Li, and
  Yuan}]{song2021stacked}
Song, L.; Wu, J.; Yang, M.; Zhang, Q.; Li, Y.; and Yuan, J. 2021.
\newblock Stacked homography transformations for multi-view pedestrian
  detection.
\newblock In \emph{Proceedings of the IEEE/CVF International Conference on
  Computer Vision}, 6049--6057.

\bibitem[{Taj and Cavallaro(2011)}]{taj2011distributed}
Taj, M.; and Cavallaro, A. 2011.
\newblock Distributed and decentralized multicamera tracking.
\newblock \emph{IEEE Signal Processing Magazine}, 28(3): 46--58.

\bibitem[{Tzeng et~al.(2017)Tzeng, Hoffman, Saenko, and
  Darrell}]{tzeng2017adversarial}
Tzeng, E.; Hoffman, J.; Saenko, K.; and Darrell, T. 2017.
\newblock Adversarial discriminative domain adaptation.
\newblock In \emph{Proceedings of the IEEE conference on computer vision and
  pattern recognition}, 7167--7176.

\bibitem[{Viola and Jones(2004)}]{viola2004robust}
Viola, P.; and Jones, M.~J. 2004.
\newblock Robust real-time face detection.
\newblock \emph{International journal of computer vision}, 57(2): 137--154.

\bibitem[{Viola, Jones, and Snow(2005)}]{viola2005detecting}
Viola, P.; Jones, M.~J.; and Snow, D. 2005.
\newblock Detecting pedestrians using patterns of motion and appearance.
\newblock \emph{International Journal of Computer Vision}, 63(2): 153--161.

\bibitem[{Wang and Zou(2016)}]{wang2016fast}
Wang, Y.; and Zou, Y. 2016.
\newblock Fast visual object counting via example-based density estimation.
\newblock In \emph{IEEE International Conference on Image Processing (ICIP)},
  3653--3657. IEEE.

\bibitem[{Wu and Nevatia(2007)}]{wu2007detection}
Wu, B.; and Nevatia, R. 2007.
\newblock Detection and tracking of multiple, partially occluded humans by
  bayesian combination of edgelet based part detectors.
\newblock \emph{International Journal of Computer Vision}, 75(2): 247--266.

\bibitem[{Xu and Qiu(2016)}]{xu2016crowd}
Xu, B.; and Qiu, G. 2016.
\newblock Crowd density estimation based on rich features and random projection
  forest.
\newblock In \emph{IEEE Winter Conference on Applications of Computer Vision
  (WACV)}, 1--8. IEEE.

\bibitem[{Xu et~al.(2016)Xu, Liu, Liu, and Zhu}]{xu2016multi}
Xu, Y.; Liu, X.; Liu, Y.; and Zhu, S.~C. 2016.
\newblock Multi-view People Tracking via Hierarchical Trajectory Composition.
\newblock In \emph{Computer Vision and Pattern Recognition}, 4256--4265.

\bibitem[{Yang et~al.(2022)Yang, Ding, Li, and Hu}]{yang2022distributed}
Yang, S.; Ding, F.; Li, P.; and Hu, S. 2022.
\newblock Distributed multi-camera multi-target association for real-time
  tracking.
\newblock \emph{Scientific Reports}, 12(1): 11052.

\bibitem[{You and Jiang(2020)}]{you2020real}
You, Q.; and Jiang, H. 2020.
\newblock Real-time 3d deep multi-camera tracking.
\newblock \emph{arXiv preprint arXiv:2003.11753}.

\bibitem[{Zhai et~al.(2022)Zhai, Yang, Li, Xie, Cheng, and Liu}]{zhai2022co}
Zhai, Q.; Yang, F.; Li, X.; Xie, G.-S.; Cheng, H.; and Liu, Z. 2022.
\newblock Co-Communication Graph Convolutional Network for Multi-View Crowd
  Counting.
\newblock \emph{IEEE Transactions on Multimedia}.

\bibitem[{Zhang et~al.(2022)Zhang, Cheng, Wu, Li, and Qiao}]{zhang2022crossnet}
Zhang, J.; Cheng, Z.-Q.; Wu, X.; Li, W.; and Qiao, J.-J. 2022.
\newblock Crossnet: Boosting crowd counting with localization.
\newblock In \emph{Proceedings of the 30th ACM International Conference on
  Multimedia}, 6436--6444.

\bibitem[{Zhang and Chan(2019)}]{zhang2019wide}
Zhang, Q.; and Chan, A.~B. 2019.
\newblock Wide-Area Crowd Counting via Ground-Plane Density Maps and Multi-View
  Fusion CNNs.
\newblock In \emph{Proceedings of the IEEE Conference on Computer Vision and
  Pattern Recognition}, 8297--8306.

\bibitem[{Zhang and Chan(2020)}]{zhang2020_3d}
Zhang, Q.; and Chan, A.~B. 2020.
\newblock 3D Crowd Counting via Multi-View Fusion with 3D Gaussian Kernels.
\newblock In \emph{AAAI Conference on Artificial Intelligence}.

\bibitem[{Zhang and Chan(2022{\natexlab{a}})}]{zhang2022_3d}
Zhang, Q.; and Chan, A.~B. 2022{\natexlab{a}}.
\newblock 3D Crowd Counting via Geometric Attention-Guided Multi-view Fusion.
\newblock \emph{International Journal of Computer Vision}, 130(12): 3123--3139.

\bibitem[{Zhang and Chan(2022{\natexlab{b}})}]{zhang2022wide}
Zhang, Q.; and Chan, A.~B. 2022{\natexlab{b}}.
\newblock Wide-area crowd counting: Multi-view fusion networks for counting in
  large scenes.
\newblock \emph{International Journal of Computer Vision}, 130(8): 1938--1960.

\bibitem[{Zhang, Lin, and Chan(2021)}]{zhang2021cross}
Zhang, Q.; Lin, W.; and Chan, A.~B. 2021.
\newblock Cross-View Cross-Scene Multi-View Crowd Counting.
\newblock In \emph{Proceedings of the IEEE/CVF Conference on Computer Vision
  and Pattern Recognition}, 557--567.

\bibitem[{Zheng, Li, and Mu(2021)}]{zheng2021learning}
Zheng, L.; Li, Y.; and Mu, Y. 2021.
\newblock Learning Factorized Cross-View Fusion for Multi-View Crowd Counting.
\newblock In \emph{2021 IEEE International Conference on Multimedia and Expo
  (ICME)}, 1--6. IEEE.

\bibitem[{Zhu et~al.(2020)Zhu, Su, Lu, Li, Wang, and Dai}]{zhu2020deformable}
Zhu, X.; Su, W.; Lu, L.; Li, B.; Wang, X.; and Dai, J. 2020.
\newblock Deformable detr: Deformable transformers for end-to-end object
  detection.
\newblock \emph{arXiv preprint arXiv:2010.04159}.

\end{thebibliography}

\end{document}